# A Universal and Robust Framework for Multiple Gas Recognition Based-on Spherical Normalization-Coupled Mahalanobis Algorithm

Shuai Chen[a], Yang Song[b], Chen Wang[a], Ziran Wang[a,c,*]

[a] *Key Laboratory of High Efficiency and Clean Mechanical Manufacture, Ministry of Education, School of Mechanical Engineering, Shandong University, Jinan, 250000, China*

[b] *Department of Mechanical Engineering, Hong Kong Polytechnic University, Hong Kong, SAR, China*

[c] *Suzhou Research Institute of Shandong University, Suzhou, 215123, China*

## Abstract

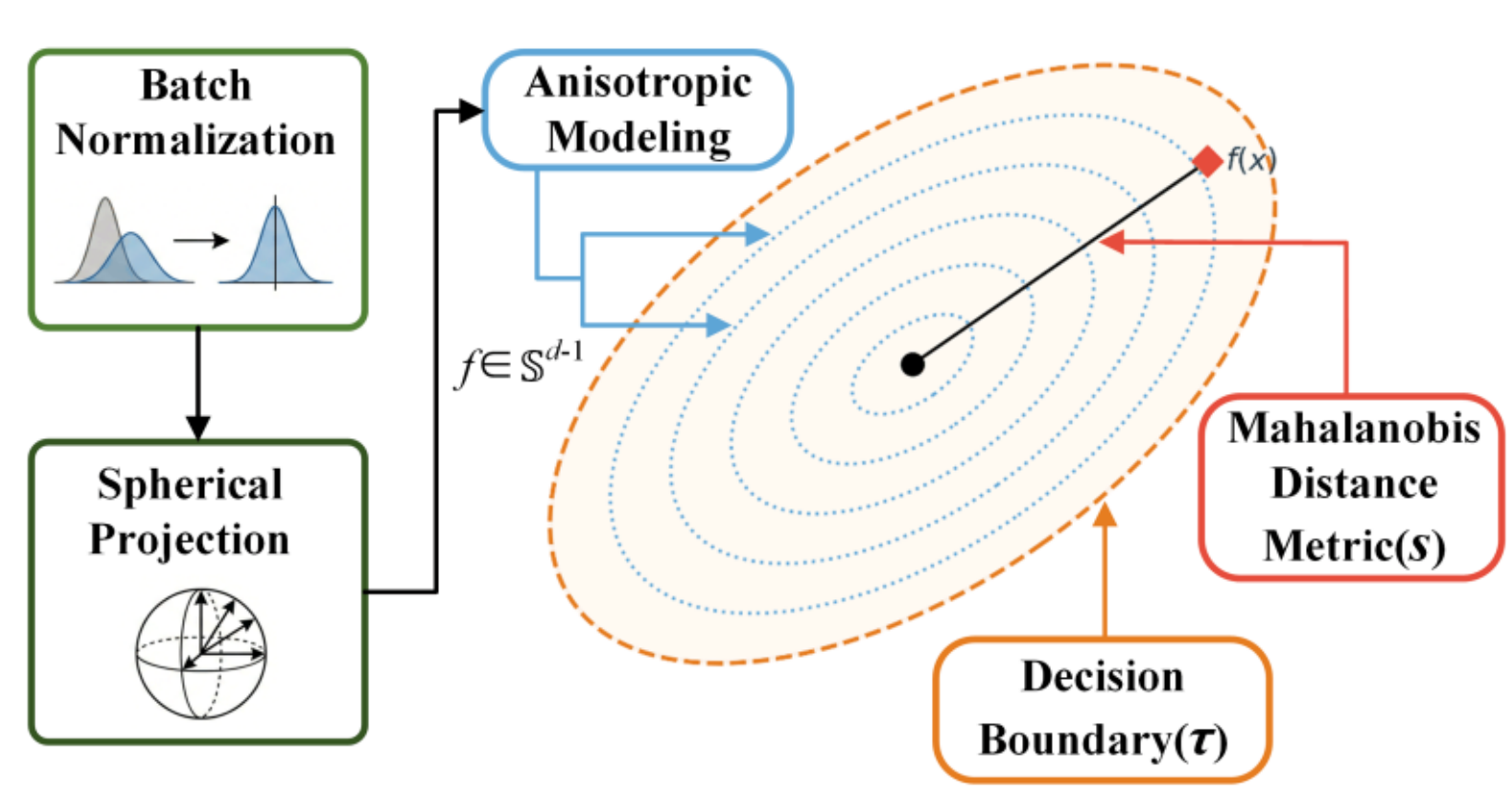
Batch Normalization
Anisotropic Modeling
f(x)
$f \in \mathbb{S}^{d-1}$
Spherical Projection
Mahalanobis Distance Metric(s)
Decision Boundary(τ)

Electronic nose (E-nose) systems face two interconnected challenges in open-set gas recognition: feature distribution shift caused by signal drift and decision boundary failure induced by unknown gas interference. Existing methods predominantly rely on Euclidean distance or conventional classifiers, failing to account for anisotropic feature distributions and dynamic signal intensity variations. To address these issues, this study proposes the Spherical Normalization coupled Mahalanobis (SNM) module, a universal post-processing module for open-set gas recognition. First, it achieves geometric decoupling through cascaded batch and L2 normalization, projecting features onto a unit hypersphere to eliminate signal intensity fluctuations. Second, it utilizes Mahalanobis distance to construct adaptive ellipsoidal decision boundaries that conform to the anisotropic feature geometry. The architecture-agnostic SNM-Module seamlessly integrates with mainstream backbones including Convolutional Neural Network (CNN), Recurrent Neural Network (RNN), and Transformer. Experiments on the public Vergara dataset demonstrate that the Transformer+SNM configuration achieves near-theoretical-limit performance in discriminating among multiple target gases, with an AUROC of 0.9977±0.0028 and an unknown gas detection rate of 99.57% at 5% false positive rate, significantly outperforming state-of-the-art methods with a 3.0% AUROC improvement and 91.0% standard deviation reduction compared to Class Anchor Clustering (CAC). The module maintains exceptional robustness across five sensor positions, with standard deviations below 0.0028. This work effectively addresses the critical challenge of simultaneously achieving high accuracy and high stability in open-set gas recognition, providing solid support for industrial E-nose deployment.

## 1. Introduction

Electronic nose (E-nose) systems, as intelligent sensing technologies that emulate biological olfactory mechanisms, respond to gas mixtures through gas-sensitive sensor arrays and achieve automatic gas identification and concentration estimation through pattern recognition algorithms.[1,2] In recent years, this technology has demonstrated promising application prospects in food quality monitoring, [3–6] medical diagnosis,[7,8] industrial safety detection,[9–11] and environmental monitoring.[12,13] With the rapid advancement of deep learning, researchers have proposed various neural network-based gas recognition methods, including convolutional neural networks,[14–16] recurrent neural networks,[17,18] and hybrid deep network architectures,[19,20] which have achieved remarkable performance in closed-set recognition scenarios.

However, when extending E-nose systems from laboratory environments to practical industrial settings, a fundamental contradiction emerges between the closed-set assumption and open-world conditions. Traditional gas recognition methods adopt the closed-set assumption, wherein test samples must belong to one of the known classes defined during training. Nevertheless, the gas composition in real industrial environments is complex and variable. Metal oxide semiconductor (MOS) sensors exhibit high sensitivity to non-target odor interference in the environment, such as perfume, alcohol, and fruit aromas. [21] These background odors interfere with normal sensor responses, leading to misidentification of target gases and biased concentration estimates. In practical applications, there exist thousands of unknown “non-target” interference sources, and effectively counteracting these unknown interferences has become a critical challenge for the practical deployment of E-nose technology.[22,23]

Open-set recognition (OSR) provides an effective approach to address the aforementioned problems. Unlike closed-set recognition, OSR requires models not only to accurately classify known classes encountered during training but also to detect and reject unknown samples that were never present during training.[24,25] Scheirer et al. first systematically defined the open-set recognition problem and proposed a solution based on the 1-vs-Set machine.[26] Subsequently, Bendale and Boult proposed the OpenMax method,[27] which models the activation vector distribution at class boundaries through extreme value theory, achieving open-set extension of deep neural networks. Rudd et al.[28] further developed the Extreme Value Machine (EVM), utilizing Weibull distributions to model sample inclusion probabilities. More recently, Miller et al. proposed the Class Anchor Clustering (CAC) method, [29] which enhances intra-class feature compactness through anchor constraint loss, achieving excellent performance on multiple benchmark datasets. Qu et al. first applied open-set recognition methods to E-nose datasets, validating the effectiveness of the CAC-CNN method in gas open-set recognition tasks, [30] achieving an average AUROC of 0.956. Building upon this, Ma et al. proposed a multi-scale temporal convolutional network integrated with an SE-ResNet channel attention mechanism (MSE-TCN) for the open-set gas classification task in electronic noses, achieving a mean AUROC of 0.9657.[31]

Despite the significant progress achieved by the aforementioned methods in the field of open-set recognition, two fundamental challenges remain when applying them to gas recognition tasks. A key limitation arises from the predominant use of Euclidean distance in existing open-set methods to construct decision boundaries, which implicitly assumes isotropic distribution in the feature space. However, the response characteristics of gas-sensitive sensor arrays exhibit significant anisotropic properties: different sensor channels show vastly different sensitivities to different gases. For instance, the TGS2611 sensor is sensitive to methane while the TGS2602 sensor is sensitive to ammonia.[32] Moreover, due to the physical proximity of sensors, non-trivial correlations exist among channels.[33,34] Consequently, the spherical decision boundaries corresponding to Euclidean distance cannot effectively adapt to such complex feature geometry, resulting in systematic deviations between boundaries and the true data manifold.

Further complicating this scenario is the issue of signal drift, which severely constrains the practical deployment capability of gas recognition systems. The response characteristics of gas-sensitive sensors are influenced by multiple factors, including dynamic fluctuations in gas concentration, variations in environmental temperature and humidity, sensor aging and poisoning effects, and spatial relationships between sensors and gas sources.[35–37] Zhang and Zhang pointed out that sensor drift has been a long-standing challenge in the E-nose field.[38] Gas samples collected by drifted sensors exhibit different distribution characteristics compared to drift-free data, making classification tasks more difficult. Multiple studies have been devoted to developing drift compensation methods,[39–42] including transfer learning, domain adaptation, and ensemble learning strategies. However, these methods typically require labeled data from the target domain or complex online update mechanisms. More critically, signal drift not only affects the recognition accuracy of known classes but also interferes with threshold settings for unknown detection. Thresholds trained under one condition may completely fail under other conditions, posing severe challenges for the practical deployment of traditional open-set methods.

Furthermore, feature vectors output by existing feature extraction methods simultaneously encodes both chemical property information and signal intensity information of gases, lacking effective mechanisms to decouple these two types of information.[43,44] When signal drift occurs, the magnitude of feature vectors changes. Even if their direction remains stable, distance-based classifiers will produce erroneous judgments. Existing methods fail to fundamentally address this issue, making it difficult for models to distinguish between genuine chemical differences and intensity variations caused by concentration fluctuations or equipment aging.

To address these fundamental contradictions, this study proposes A universal and robust framework designed to reconstruct the feature space geometry for the precise recognition of multiple gas species. (Figure 1) The proposed approach fundamentally reshapes the decision logic by coupling Spherical Normalization with Mahalanobis distance (SNM) module. By projecting deep features onto a unit hypersphere via cascaded batch and L2 normalization, the module mathematically achieves a complete decoupling of feature direction from magnitude. This transformation compels the model to rely solely on directional information representing chemical properties, thereby eliminating interference from physical signal intensity fluctuations at the root. In conjunction with this geometric projection, Mahalanobis distance is utilized as the core scoring mechanism. By modeling correlations among feature dimensions through covariance matrices, it constructs adaptive ellipsoidal decision boundaries that achieve conformal adaptation to the true anisotropic data manifold. Designed as an architecture-agnostic post-processing unit, the SNM-Module can be seamlessly embedded into mainstream backbone networks such as CNN, RNN, and Transformer, elevating the performance of identifying diverse gas targets in open-set scenarios to near-theoretical limits.

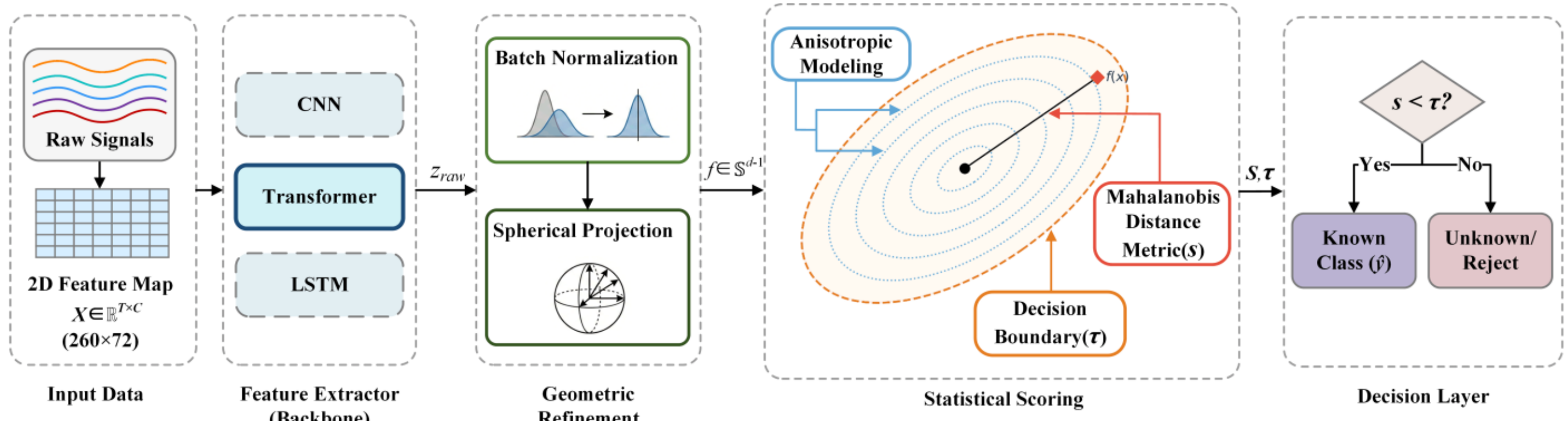

**Figure 1.** Overall architecture diagram of the SNM-Module.

## 2. Materials and Methods

### 2.1 Dataset Description

This study employed the publicly available gas sensor dataset released by Vergara et al. for experimental validation. This dataset enjoys widespread recognition in the gas recognition field and has been used to evaluate the performance of various recognition algorithms. Data collection was conducted in a wind tunnel environment, yielding 18,000 time-series gas samples covering ten high-priority chemical gaseous substances: acetone, acetaldehyde, ammonia, butanol, ethylene, methane, methanol, carbon monoxide, benzene, and toluene.

The experimental system utilized nine identical metal oxide gas sensor modules, each containing eight different commercial sensors (TGS series), totaling 72 sensing elements. Table 1 provides detailed specifications of each sensor model and their target detection gases. This redundant design enhances system robustness against sensor noise and failures while providing rich multidimensional feature information. Release concentrations for each gas were set according to their physicochemical properties and practical application scenarios, with source concentrations ranging from 100 ppmv for butanol to 10,000 ppmv for ammonia.

It should be noted that this study focuses on addressing signal drift within a single sensor array system. The nine sensor modules employed in this dataset are identical in hardware configuration, allowing us to isolate and study drift-induced distribution shifts caused by environmental factors such as distance, concentration decay, and temporal variations. Cross-device generalization, which involves transferring models across different sensor arrays with potential hardware variations and manufacturing inconsistencies, represents an important but distinct research direction beyond the scope of this work. The proposed SNM-Module primarily targets drift-induced distribution shifts within the same physical system, which constitutes a critical prerequisite for practical E-nose deployment in real-world scenarios.

Data acquisition for each sample lasted 260 seconds, during which sensor resistance values were recorded at a frequency of 100 Hz. To systematically evaluate the impact of signal drift on recognition performance, data collection was conducted at six different positions (L1-L6), with distances between the sensor array and gas release source incrementally increasing from 1 to 4 meters. As distance increases, gas concentration decays exponentially, resulting in significantly reduced sensor response intensity. This study utilized data from positions L1-L5. Position L6 was excluded primarily due to incomplete raw data records that resulted in format inconsistencies incompatible with the standardized preprocessing pipeline. Additionally, L6 represents the most extreme drift condition at 4 meters from the gas source, where sensor responses approached noise floor levels. The L1-L5 range already encompasses substantial drift variations with an approximate tenfold concentration decay, providing sufficient challenge for evaluating drift robustness while maintaining practical relevance for real-world deployment scenarios. Each sub-dataset contains 300 samples per gas class.

Raw sensor response sequences underwent the following preprocessing steps. Temporal downsampling was initially performed by computing the average of 100 sampling points per second, reducing the sampling frequency to 1 Hz and thereby regularizing the response sequence to 260 time steps. Subsequently, z-score normalization was applied channel-wise to eliminate baseline differences among different sensors. The processed data was then restructured into a 260×72 two-dimensional spatiotemporal feature map, where 260 represents the temporal dimension and 72 represents the spatial channel dimension.

**Table 1.** Sensor configuration of the Vergara gas sensing dataset

| Sensor Model | Quantity per Array | Target Gases (Manufacturer Specifications) |
|---|---|---|
| TGS2600 | 1 | Hydrogen, Carbon monoxide |
| TGS2602 | 2 | Ammonia, Hydrogen sulfide, VOCs |
| TGS2610 | 1 | Propane |
| TGS2611 | 1 | Methane |
| TGS2612 | 1 | Methane, Propane, Butane |
| TGS2620 | 2 | Carbon monoxide, Combustible gases, VOCs |
| **Total** | **8** | (9 modules × 8 sensors = 72 sensors) |

## 2.2 Experimental Protocol

This study employed a 10-fold cross-validation strategy to evaluate model performance. For each sub-dataset, the ten gas types were randomly partitioned into six known classes and four unknown classes. Sixty percent of samples from known classes were randomly selected as the training set (1,080 samples in total), while samples from unknown classes were excluded from training. The test set comprised 40% of randomly selected samples from unknown classes combined with remaining samples from known classes (1,200 samples in total). This random partitioning process was repeated 10 times to evaluate each model.

To comprehensively assess model performance under different conditions, this study designed 50 independent experiments (5 positions × 10 folds). Each experiment conducted a complete training-validation-testing procedure on a specific position and fold partition, with final results reported as statistical summaries across all experiments (mean ± standard deviation). This design ensures both evaluation comprehensiveness and adequate sample size for statistical significance analysis.

This study employed three complementary metrics for comprehensive model performance evaluation. Known Class Accuracy measures the model's classification capability for known gas classes, calculated as the proportion of correctly classified samples among known class test samples. True Positive Rate at Fixed False Positive Rate, denoted as TPR@FPR=5%, measures the model's capability to detect unknown gases under a constraint of 5% false positive rate, directly reflecting detection efficacy in practical deployment scenarios. Area Under the ROC Curve comprehensively measures unknown detection performance across all possible thresholds and serves as the core evaluation metric for open-set recognition.

During model training, a validation set comprising 20% of the training data was employed for early stopping and hyperparameter tuning to prevent overfitting. However, for performance evaluation in open-set scenarios, AUROC was adopted as the primary metric because it is threshold-independent and provides a comprehensive performance measure across all operating points. This approach is more suitable than selecting a single fixed threshold based on validation data, as it eliminates the need for threshold optimization and better reflects the model's intrinsic capability to discriminate between known and unknown classes across diverse deployment conditions.

## 2.3 SNM-Module Framework Architecture

The SNM-Module operates as a universal post-processing module designed to decouple chemical information from signal intensity variations. As illustrated in Figure 1, the overall architecture comprises five core components: an input data processing module, a feature extractor backbone, a geometric refinement module, a statistical scoring module, and a decision layer. The input two-dimensional feature map first undergoes discriminative feature extraction through the backbone network. To address the fundamental challenge of signal drift, the subsequent processing enforces a rigorous geometric transformation detailed in Figure 2. Specifically, raw features typically exhibit systematic bias and strong intensity-class coupling (Figure 2a). The module progressively rectifies this space by recentering the distribution through batch normalization (Figure 2b) and projecting vectors onto a unit hypersphere (Figure 2c). Finally, open-set decisions are derived by the statistical scoring module based on adaptive Mahalanobis distance calculations.

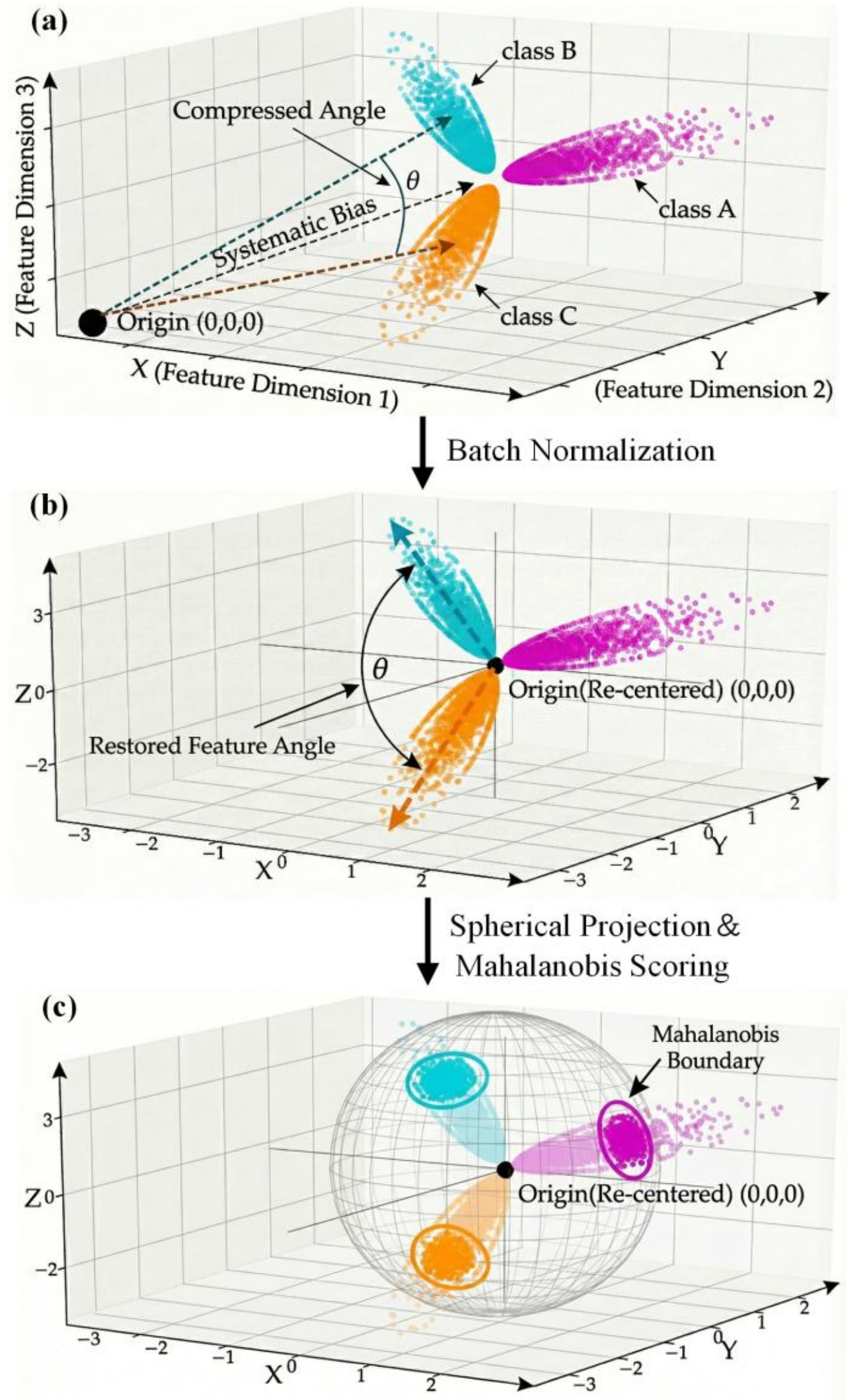

**Figure 2.** Schematic diagram of signal drift optimization. (a) Original feature space (bias and anisotropy) (b) Features after BN (centering and rescaling) (c) Features after spherical projection and Mahalanobis distance

### *2.3.1 Backbone Network Design*

The SNM-Module framework possesses architecture-agnostic properties and supports multiple types of feature extraction backbone networks. This study implemented and compared three representative architectures.

The Transformer encoder serves as the primary configuration in this study, effectively capturing long-range dependencies in spatiotemporal sequences through self-attention mechanisms. Let the input feature map be $\mathbf{X} \in \mathbb{R}^{T \times C}$, where $T = 260$ denotes the number of time steps and $C = 72$ denotes the number of channels. The input is initially mapped to the model dimension through linear projection:

$$\mathbf{H}^{(0)} = \mathbf{X}\mathbf{W}_{\text{proj}} + \mathbf{E}_{\text{pos}}$$

where $\mathbf{W}_{\text{proj}} \in \mathbb{R}^{C \times d_{\text{model}}}$ is the projection matrix, $d_{\text{model}} = 128$ is the model dimension, and $\mathbf{E}_{\text{pos}} \in \mathbb{R}^{T \times d_{\text{model}}}$ represents learnable positional encodings.

The Transformer encoder consists of $L = 2$ stacked layers, each containing a multi-head self-attention sublayer and a feed-forward network sublayer. The multi-head attention mechanism is defined as:

$$\text{MultiHead}(\mathbf{H}) = \text{Concat}(\text{head}_1, \ldots, \text{head}_h)\mathbf{W}^O$$

$$\text{head}_i = \text{Attention}\left(\mathbf{H}\mathbf{W}_i^Q, \mathbf{H}\mathbf{W}_i^K, \mathbf{H}\mathbf{W}_i^V\right)$$

This study sets the number of attention heads to $h = 4$. Finally, temporal information is aggregated through global average pooling to obtain a fixed-length feature vector $\mathbf{z}_{\text{raw}} \in \mathbb{R}^d$.

For comparison, this study also implemented a CNN backbone network and a bidirectional LSTM backbone network. The CNN architecture is well-suited for gas recognition as it can extract multi-scale spatial patterns from the sensor array responses. Through hierarchical convolutions, CNN captures both local sensor-specific features and global cross-sensor correlations, which are critical for distinguishing different gas signatures. The bidirectional LSTM is particularly effective for processing gas sensor time series, as it can model the temporal dynamics of gas adsorption and desorption processes. By processing the sequence in both forward and backward directions, LSTM captures the characteristic rising and falling phases of sensor responses through its gating mechanisms, enabling robust temporal feature learning.

#### *2.3.2 Geometric Refinement Module*

The geometric refinement module represents one of the core innovations of the SNM-Module. Through cascaded batch normalization and L2 normalization, it projects raw features onto the unit hypersphere, achieving decoupling of feature direction and magnitude.

Let the raw features output by the backbone network be $\mathbf{z}_{\text{raw}} \in \mathbb{R}^d$. The batch normalization operation is defined as:

$$\hat{\mathbf{z}} = \frac{\mathbf{z}_{\text{raw}} - \boldsymbol{\mu}_B}{\sqrt{\boldsymbol{\sigma}_B^2 + \epsilon}}$$

where $\boldsymbol{\mu}_B$ and $\boldsymbol{\sigma}_B^2$ are the mean and variance vectors of the current batch, respectively, and $\epsilon$ is a numerical stability constant. Batch normalization eliminates distributional shift and scale differences in features, providing standardized distributions with an approximate zero mean and unit variance, thereby establishing a stable input for subsequent spherical projection.

L2 normalization is then applied to batch-normalized features, projecting them onto the unit hypersphere $\mathbb{S}^{d-1}$:

$$\mathbf{f} = \frac{\hat{\mathbf{z}}}{\| \hat{\mathbf{z}} \|_2}$$

where $\|\cdot\|_2$ denotes the Euclidean norm. The projected features $\mathbf{f}$ satisfy the constraint $\| \mathbf{f} \|_2 = 1$, meaning all feature vectors reside on the $(d-1)$-dimensional unit hypersphere. This geometric refinement process is based on a simplifying assumption that raw feature vectors can be approximately decomposed into a directional component represented by the unit vector $\mathbf{f}$ and a magnitude component represented by the scalar $\| \mathbf{z}_{raw} \|_2$. Under this first-order approximation, the directional component primarily encodes chemical property information of gases, while the magnitude component primarily reflects signal intensity information. While the actual relationship between chemical properties and signal intensity may involve nonlinear coupling effects, our experimental results in Section 3.4 demonstrate that this geometric decomposition is effective in practice. Through spherical projection, model decisions rely primarily on directional information, substantially reducing interference from signal intensity fluctuations across different drift conditions.

Figure 3 demonstrate the effects of geometric refinement. Raw feature magnitudes exhibit significant differences across sensor positions, whereas after geometric refinement, feature magnitudes from all positions are uniformly normalized to 1.0, achieving cross-position geometric alignment.

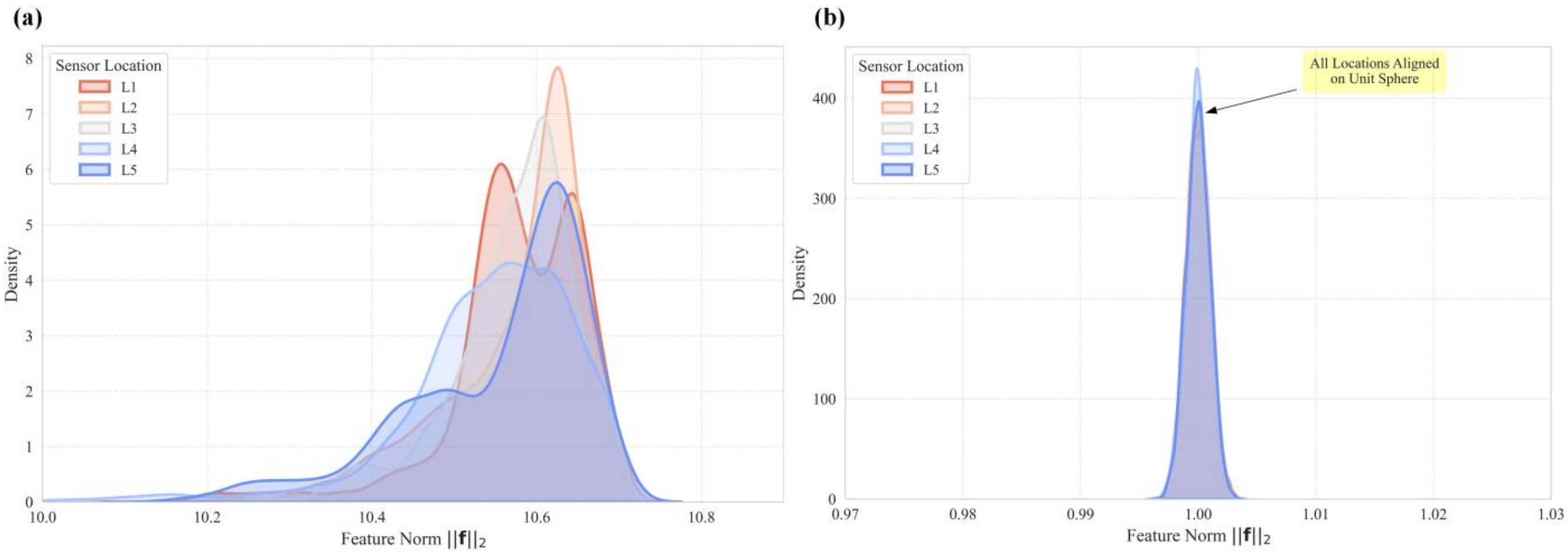

**Figure 3.** The effects of geometric refinement. (a) Feature magnitude distribution across different positions (L1-L5) in the original feature space (b) Features after spherical normalization with all positions aligned

### *2.3.3 Statistical Scoring Module*

The statistical scoring module calculates confidence scores for sample membership in each known class based on Mahalanobis distance and subsequently makes open-set decisions. During training, statistics are computed for each known class $c \in \{1,2,\dots,K\}$:

$$\boldsymbol{\mu}_c = \frac{1}{N_c} \sum_{i:y_i=c} \mathbf{f}_i$$

$$\boldsymbol{\Sigma}_c = \frac{1}{N_c - 1} \sum_{i:y_i=c} (\mathbf{f}_i - \boldsymbol{\mu}_c)(\mathbf{f}_i - \boldsymbol{\mu}_c)^\top$$

where $\boldsymbol{\mu}_c \in \mathbb{R}^d$ is the mean vector (class center) of class $c$, $\boldsymbol{\Sigma}_c \in \mathbb{R}^{d \times d}$ is the covariance matrix of class $c$, and $N_c$ is the number of samples in class $c$. To improve numerical stability and prevent covariance matrix singularity, a regularization term is introduced:

$$\widetilde{\boldsymbol{\Sigma}}_c = \boldsymbol{\Sigma}_c + \lambda \mathbf{I}$$

where $\lambda = 10^{-4}$ is the regularization coefficient and $\mathbf{I}$ is the identity matrix.

For a test sample $\mathbf{x}$, the Mahalanobis distance from its spherical feature $\mathbf{f}(\mathbf{x})$ to each class center is computed:

$$D_M(\mathbf{f}(\mathbf{x}), c) = \sqrt{(\mathbf{f}(\mathbf{x}) - \boldsymbol{\mu}_c)^\top \widetilde{\boldsymbol{\Sigma}}_c^{-1} (\mathbf{f}(\mathbf{x}) - \boldsymbol{\mu}_c)}$$

The fundamental advantage of Mahalanobis distance over Euclidean distance lies in: automatically weighting different feature dimensions through the inverse covariance matrix $\widetilde{\boldsymbol{\Sigma}}_c^{-1}$ while accounting for inter-dimensional correlations. As illustrated in Figure 4, this results in ellipsoidal decision boundaries that adaptively conform to the anisotropic geometric structure of data distributions. In practical computation, Cholesky decomposition is employed to enhance numerical stability and computational efficiency.

The rejection score for sample $\mathbf{x}$ is defined as its Mahalanobis distance to the nearest class center:

$$s(\mathbf{x}) = \min_{c \in \{1,\ldots,K\}} D_M(\mathbf{f}(\mathbf{x}), c)$$

The predicted class is the one with the minimum distance:

$$\hat{y}(\mathbf{x}) = \operatorname{argmin}_{c \in \{1,\ldots,K\}} D_M(\mathbf{f}(\mathbf{x}), c)$$

The final decision rule is:

$$\text{Output} = \begin{cases} \hat{y}(\mathbf{x}) & \text{if } s(\mathbf{x}) < \tau \\ \text{Unknown (Reject)} & \text{if } s(\mathbf{x}) \geq \tau \end{cases}$$

where $\tau$ is the rejection threshold, adaptively determined based on the 95th percentile of the score distribution on the validation set.

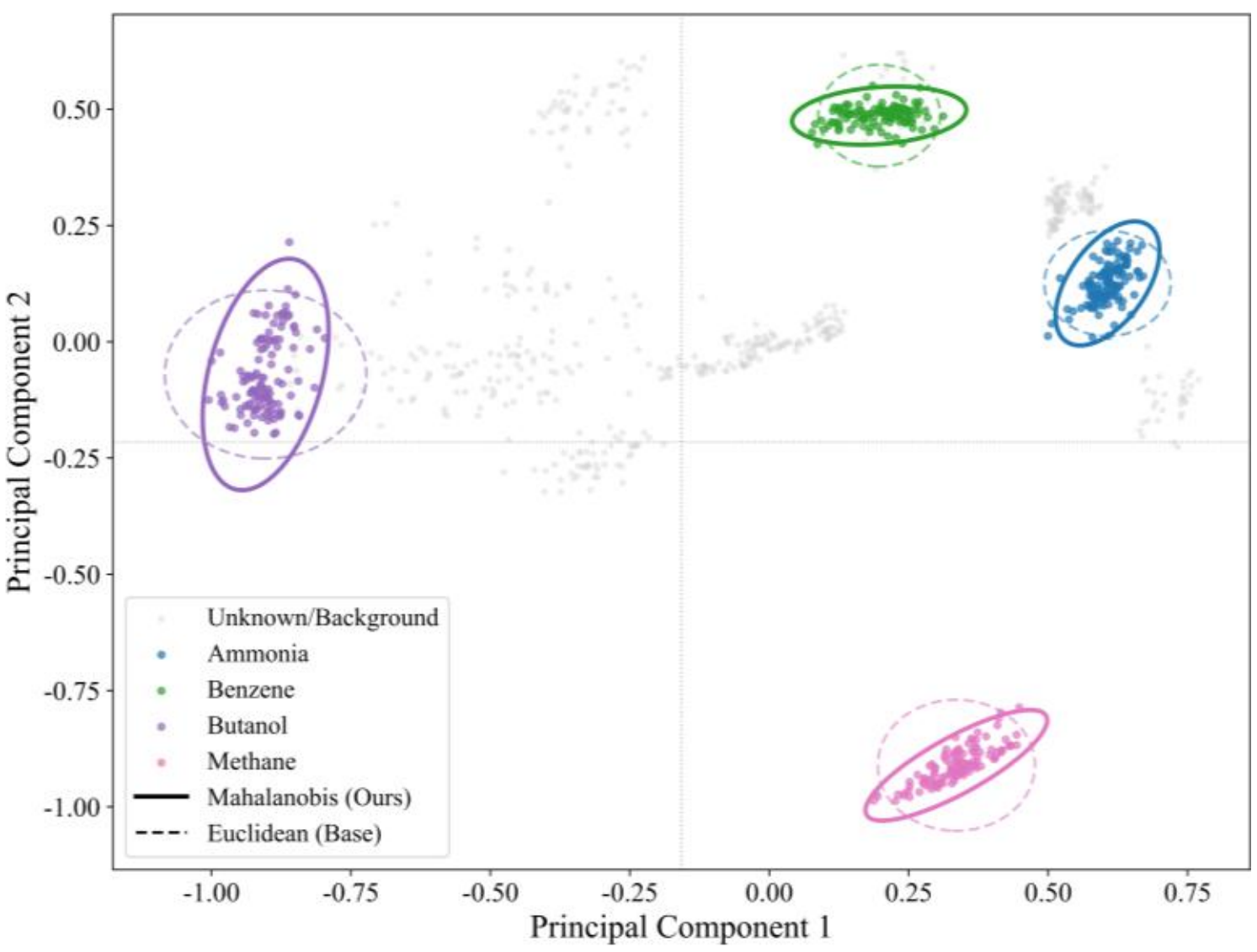

**Figure 4.** Comparison of Mahalanobis distance. (solid ellipse) and Euclidean distance (dashed circle) decision boundaries

## 2.4 Training Strategy and Hyperparameters

Table 2 provides detailed hyperparameter configurations for model training. This study employs the CAC loss function for training, which simultaneously optimizes classification performance and intra-class feature compactness:

$$\mathcal{L} = \mathcal{L}_{\text{CE}} + \lambda_{\text{anchor}} \mathcal{L}_{\text{anchor}}$$

where $\mathcal{L}_{\text{CE}}$ is the cross-entropy loss, $\mathcal{L}_{\text{anchor}}$ is the anchor constraint loss, and $\lambda_{\text{anchor}} = 10^{-5}$ is the balancing coefficient. The Adam optimizer is employed with an initial learning rate of $4 \times 10^{-5}$ and a weight decay coefficient of $10^{-4}$. Training incorporates an early stopping mechanism: training terminates when validation loss fails to decrease for 10 consecutive epochs, and model parameters with optimal validation performance are saved.

**Table 2.** Model hyperparameter configuration

| Parameter | Value | Description |
|---|---|---|
| Optimizer | Adam | Model parameter update optimizer |
| Batch size | 16 | Samples per gradient descent iteration |
| Learning rate | $4 \times 10^{-5}$ | Initial optimizer learning rate |
| Weight decay | $1 \times 10^{-4}$ | L2 regularization coefficient |

| Parameter | Value | Description |
|---|---|---|
| Max epochs | 25 | Maximum training iterations |
| Early stopping patience | 10 | Tolerance epochs without validation loss decrease |
| Random seed | 41 | Ensures experimental reproducibility |
| Dropout rate | 0.1 | Random neuron deactivation ratio |
| $\lambda$ (CAC loss) | $1\times10^{-5}$ | Anchor constraint term weight |
| Covariance regularization | $1\times10^{-4}$ | Ensures covariance matrix positive definiteness |

### 2.5 Baseline Methods and Comparison Models

To comprehensively evaluate the performance of the SNM-Module, this study implemented the following comparison methods. The Softmax baseline trains a classifier using standard cross-entropy loss and uses the negative maximum Softmax probability as the rejection score, representing the direct application of traditional closed-set classifiers to open-set scenarios. EVM constructs decision boundaries based on extreme value theory, modeling class boundaries through Weibull distributions, and represents a classical open-set recognition method. CAC introduces the class anchor concept and enhances intra-class feature compactness through anchor constraint loss, using Euclidean distance as the scoring mechanism, and represents a state-of-the-art method in the open-set recognition field. To ensure fair comparison, all comparison methods utilize identical backbone network architectures, training data partitions, and hyperparameter configurations.

## 3. Results and Discussion

### 3.1 Ablation Study: Verification of Individual Module Contributions

To systematically validate the individual contributions and synergistic effects of each SNM-Module component, this section conducts ablation experiments using the CNN backbone network. CNN was selected as the backbone for ablation experiments because its moderate baseline performance clearly demonstrates the gain contributed by each module. Table 3 presents quantitative results of the ablation experiments, where the baseline configuration (BASE-CNN) uses only CAC loss and Euclidean distance scoring as the control group.

**Table 3.** Quantitative results of ablation experiments (CNN backbone network)

| Configuration | Accuracy (Known) | TPR (Unknown) | AUROC |
|---|---|---|---|
| BASE-CNN (CAC only) | 0.9993±0.0024 | 0.8246±0.1492 | 0.9504±0.0474 |
| CNN+M | 0.9974±0.0062 | 0.9045±0.0837 | 0.9785±0.0192 |

| Configuration | Accuracy (Known) | TPR (Unknown) | AUROC |
|---|---|---|---|
| CNN+M+BN | 0.9991±0.0026 | 0.9347±0.0732 | 0.9847±0.0169 |
| CNN+M+L2N | 0.9986±0.0052 | 0.9270±0.0811 | 0.9829±0.0168 |
| **CNN+M+BN+L2N (Full)** | **0.9988±0.0042** | **0.9370±0.0683** | **0.9846±0.0148** |

Comparing BASE-CNN with CNN+M configurations reveals that introducing Mahalanobis distance improved AUROC from 0.9504 to 0.9785, an absolute improvement of 2.81 percentage points. TPR improved from 82.46% to 90.45%, an absolute improvement of 7.99 percentage points. Simultaneously, the standard deviation of AUROC decreased from 0.0474 to 0.0192, a reduction of 59.5%. These results demonstrate that Mahalanobis distance effectively captures the anisotropic distribution of features through covariance modeling, significantly enhancing both the accuracy and stability of unknown detection.

Further comparison between CNN+M and CNN+M+BN configurations show that introducing batch normalization improved AUROC to 0.9847, TPR to 93.47%, and reduced standard deviation from 0.0192 to 0.0169, indicating that batch normalization enhances cross-condition model consistency by stabilizing feature distributions. Similarly, L2 normalization also yields significant improvements: AUROC increased from 0.9785 to 0.9829, and standard deviation decreased from 0.0192 to 0.0168, validating the effectiveness of spherical projection in eliminating intensity information. The full configuration (CNN+M+BN+L2N) achieves the lowest standard deviation (0.0148), demonstrating that the cascade of BN and L2N indeed enhances model robustness. Compared to the baseline, AUROC improved by 3.42 percentage points and standard deviation decreased by 68.8%.

## 3.2 Cross-Architecture Generalization Verification

To verify the architecture-agnostic properties of the SNM-Module, this section conducts systematic comparative experiments across three representative backbone networks: CNN, LSTM, and Transformer. Figure 5 presents grouped bar charts comparing cross-architecture performance.

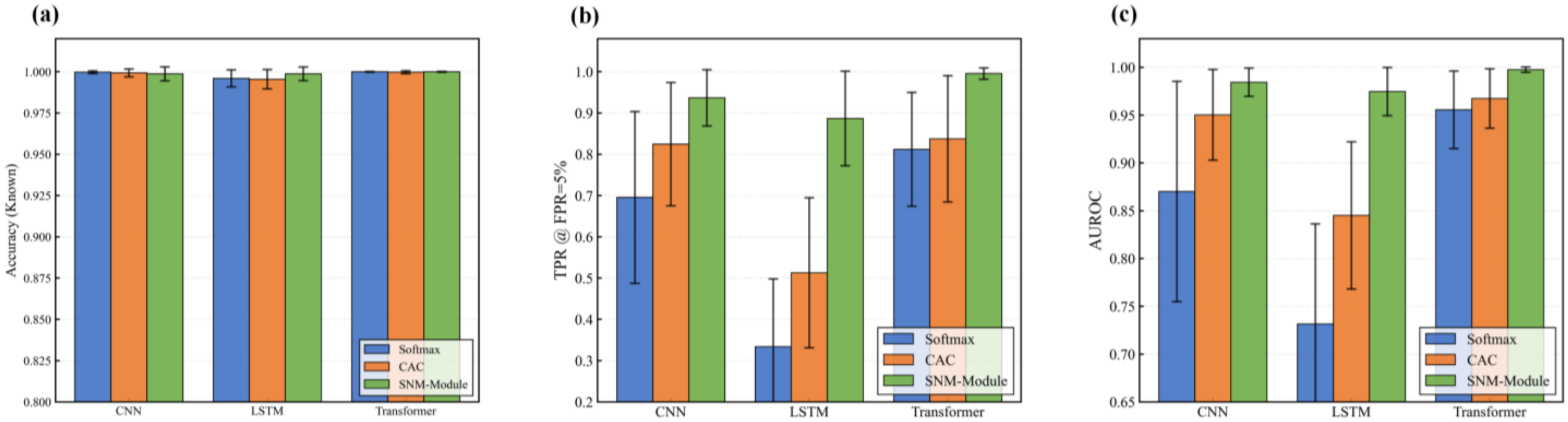

**Figure 5.** Grouped bar charts of cross-architecture performance comparison. (a) Known class classification accuracy (b) Unknown detection rate (TPR@FPR=5%) (c) Overall performance (AUROC)

For known class classification tasks, all three methods perform excellently, with accuracy generally exceeding 99.7%, indicating that the introduction of open-set mechanisms does not compromise the model's discriminative capability for known classes. Unknown detection rate serves as the key metric differentiating method performance: SNM-Module achieves significant improvements across all architectures. On the CNN backbone, SNM-Module improves over CAC by 11.24 percentage points, on the LSTM backbone by 50.40 percentage points, and on the Transformer backbone by 15.81 percentage points. Even on the less expressive LSTM backbone, SNM-Module still achieves a 95.57% unknown detection rate.

The AUROC metric comprehensively reflects model performance across all possible thresholds. SNM-Module achieves AUROC > 0.97 across all three architectures: CNN+SNM at 0.9846 ± 0.0148, LSTM+SNM at 0.9748 ± 0.0198, and Transformer+SNM at 0.9977 ± 0.0028. Experimental results demonstrate that SNM-Module provides greater absolute improvements for weaker backbones while pushing stronger backbones toward near-theoretical limits.

## 3.3 Comprehensive Comparison with State-of-the-Art Methods

Table 4 presents a comprehensive performance comparison between SNM-Module and various open-set recognition methods using the Transformer backbone network.

**Table 4.** Comprehensive performance comparison with state-of-the-art methods (Transformer backbone network)

| Method | Accuracy (Known) | TPR (Unknown) | AUROC | Remarks |
|---|---|---|---|---|
| Softmax+Transformer | 1.0000±0.0002 | 0.8120±0.1378 | 0.9557±0.0406 | |
| EVM + Transformer | 0.9999±0.0004 | 0.8227±0.1339 | 0.9514±0.0424 | |
| CAC + Transformer | 0.9998±0.0009 | 0.8376±0.1529 | 0.9675±0.0311 | |
| CAC+CNN | 0.9999 | - | 0.9560±0.0220 | Qu et al., 2022[30] |
| MSE-TCN | 0.9999 | - | 0.9657 | Ma et al., 2024[31] |
| **SNM+Transformer** | **1.0000±0.0002** | **0.9957±0.0138** | **0.9977±0.0028** | Ours |

Compared to the Softmax baseline, SNM-Module improves AUROC by 4.20 percentage points, TPR by 18.37 percentage points, and reduces standard deviation from 0.0406 to 0.0028, a decrease of 93.1%. Compared to CAC, SNM-Module improves AUROC by 3.02 percentage points, TPR by 15.81 percentage points, and reduces standard deviation by 91.0%. This significant improvement is attributed to two factors: SNM-Module's geometric refinement module eliminates intensity drift that CAC cannot handle, and Mahalanobis distance more accurately describes anisotropic feature distributions compared to CAC's Euclidean distance. Figure 6 presents ROC curve comparisons among major methods, where the Transformer-SNM curve (red solid line) is closest to the upper-left corner with an AUC of 0.9977, significantly outperforming Transformer-CAC, LSTM-SNM, and Transformer-Softmax.

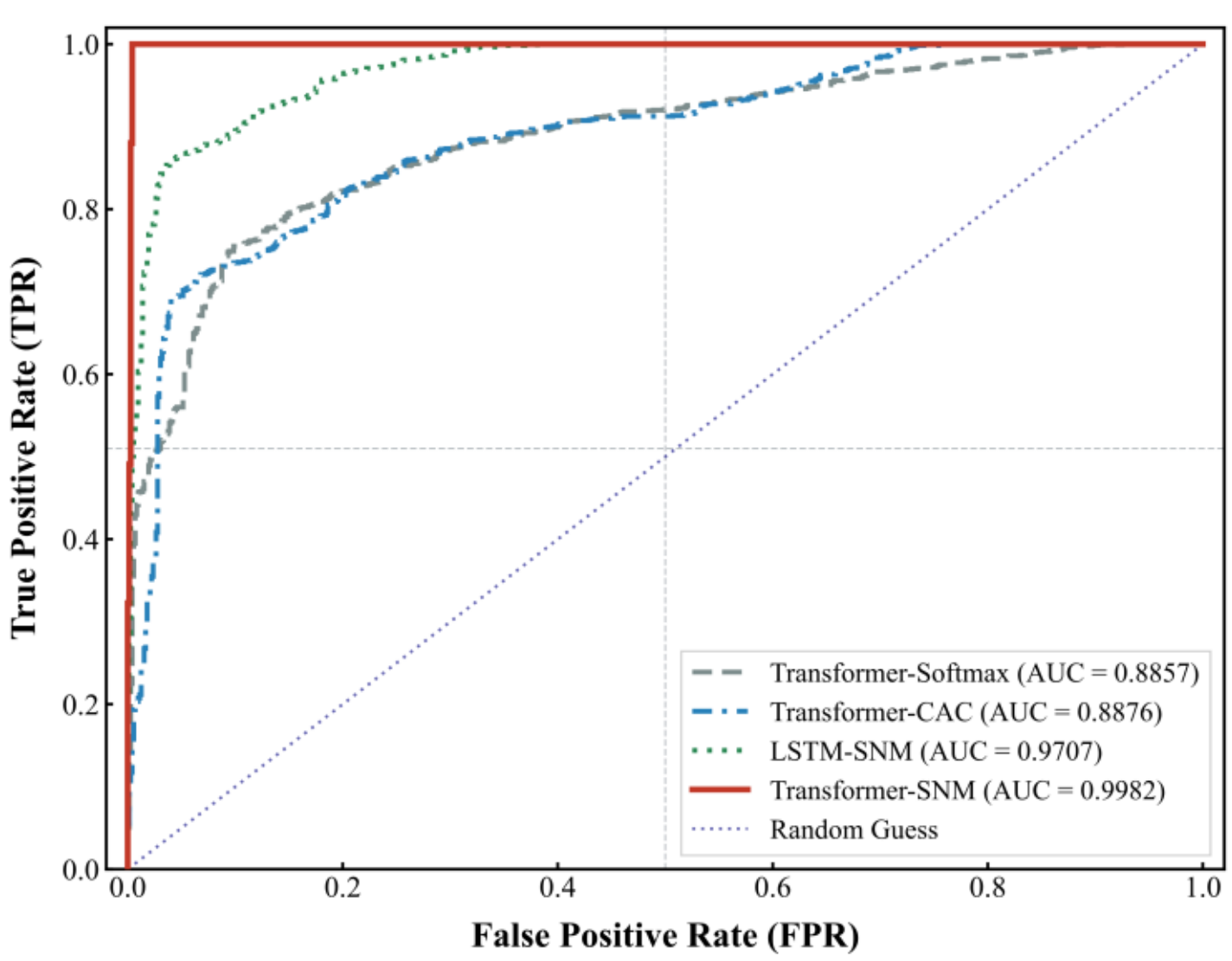

**Figure 6.** ROC curve comparison among multiple methods.

## 3.4 Multi-Position Robustness Analysis

The five positions in the Vergara dataset span distances of 1 to 4 meters from the gas release source, with signal intensity differences reaching several-fold to tens-fold. Figure 7 presents the performance of three methods across the five positions.

SNM-Module achieves the following AUROC values across the five positions: L1=0.9982, L2=0.9990, L3=0.9989, L4=0.9965, and L5=0.9959. The standard deviation of AUROC across the five positions is merely 0.0013, indicating high model insensitivity to position variations. In contrast, the Softmax baseline exhibits a standard deviation of 0.0163, which is 12.5 times that of SNM-Module. Even at position L5 with the weakest signal, SNM-Module's TPR still reaches 98.34%, while CAC achieves only 73.69%. This position invariance stems from the spherical projection eliminating the influence of magnitude, enabling model decisions to rely solely on directional information.

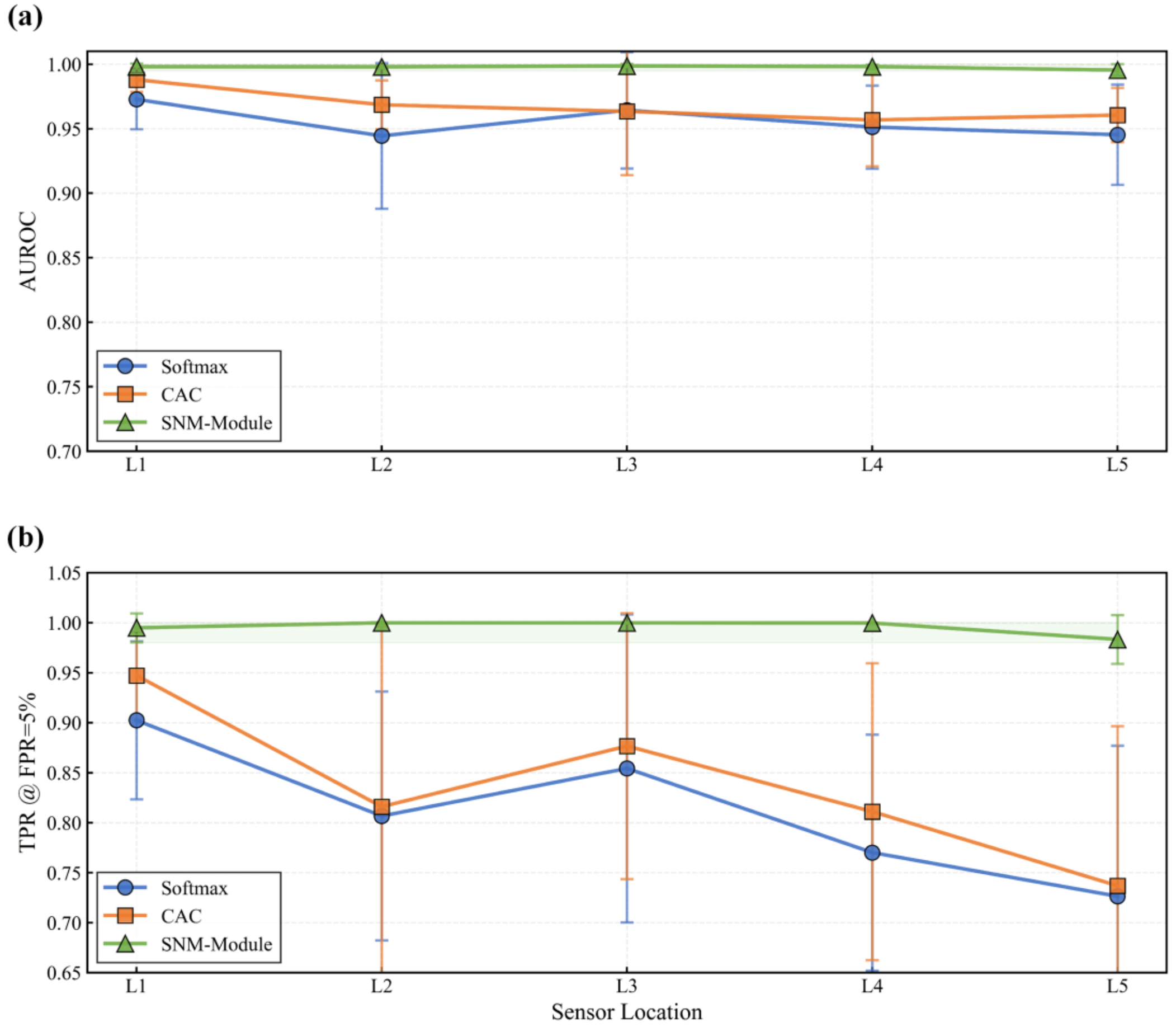

**Figure 7.** Multi-position robustness comparison charts. (a) AUROC performance across locations(b) Unknown detection rate across locations (TPR@FPR=5%)

## 3.5 Mechanistic Analysis

This section provides an in-depth analysis of the underlying mechanisms enabling SNM-Module's high performance through feature space visualization and decision space visualization. Figure 8 presents comparative visualization results between SNM-Module and the standard CAC method.

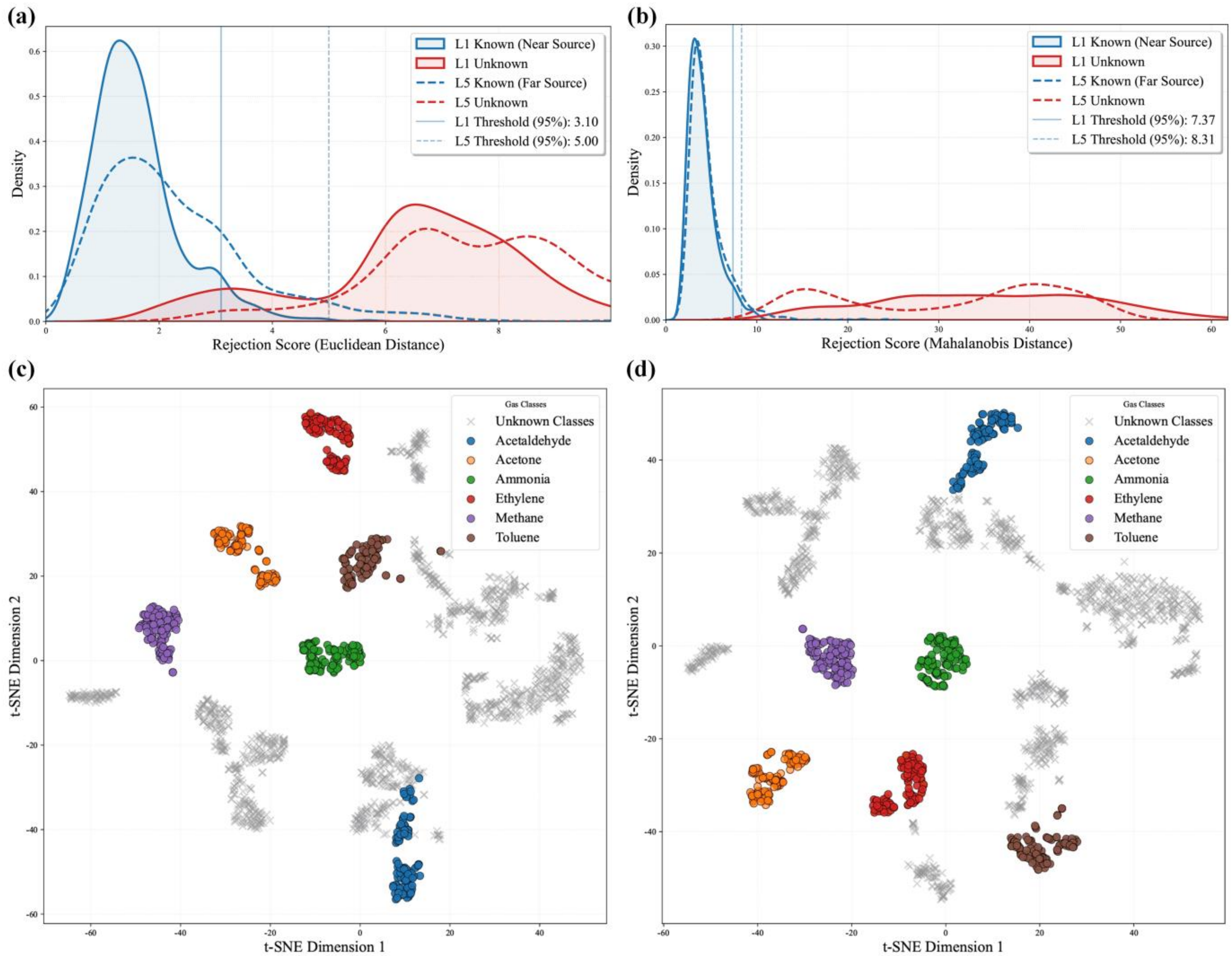

**Figure 8.** Visualization verification plots. (a) Rejection score distributions of standard CAC method at positions L1 and L5 (b) Rejection score distributions of SNM-Module at positions L1 and L5 (c) t-SNE feature space visualization of standard CAC method (d) t-SNE feature space visualization of SNM-Module

SNM-Module eliminates radial degrees of freedom at the physical level through spherical manifold projection, forcing the model to focus exclusively on directional vectors representing chemical properties.

The core of this mechanism lies in the synergistic action of cascaded BN and L2N. Batch normalization ensures input features possess stable distributions with zero mean and unit variance, providing consistent numerical conditions for L2 normalization. Subsequently, L2 normalization projects all feature vectors onto the unit hypersphere, enabling model decisions to be based entirely on directional information, decoupled from signal intensity.

This mechanism is intuitively validated in the rejection score distributions. As shown in Figure 8(a), the rejection score distributions of known classes at positions L1 and L5 under the standard CAC method exhibit clear separation, with significant differences in 95th percentile thresholds. This position dependency implies that thresholds determined through training at one position may produce excessively high false positive rates or miss rates at other positions. In contrast, as shown in Figure 8(b), the known class distributions at positions L1 and L5 after SNM-Module processing exhibit high overlap, with 95th percentile threshold differences less than 0.02 units, demonstrating that the geometric refinement module successfully eliminates distribution drift caused by sensor position differences.

The fundamental advantage of Mahalanobis distance over Euclidean distance lies in its adaptivity to the geometric structure of feature distributions. Mahalanobis distance automatically learns anisotropic structures through covariance matrices, where eigenvalues of the covariance matrix reflect variance magnitudes along principal component directions and eigenvectors define the directions of principal components. This metric assigns higher weights to directions with small variance and lower weights to directions with large variance, resulting in ellipsoidal decision boundaries that adaptively conform to the true data manifold. The t-SNE visualization results (Figure 8c-d) reveal notable differences in feature space organization. A particularly striking observation is the behavior of the blue class samples: under the standard CAC method (Figure 8c), these samples exhibit substantial overlap with unknown class regions, making discrimination challenging. In contrast, after SNM-Module processing (Figure 8d), the blue class samples are clearly separated from unknown classes and form a distinct, well-defined cluster. This visual improvement demonstrates that SNM-Module effectively enhances the discriminability between known and unknown classes in the feature space.

The combination of spherical constraints and Mahalanobis distance forms a powerful synergistic effect. Spherical projection reduces the intrinsic dimensionality of features by eliminating radial degrees of freedom, making feature distributions more regular. Mahalanobis distance then precisely captures the covariance structure of directional features on the sphere, achieving adaptive decision boundaries. From a signal processing perspective, geometric normalization eliminates the primary source of drift through amplitude variation removal, while statistical scoring handles remaining feature distribution differences in the directional domain through covariance structure modeling.

### 3.6 In-Depth Comparison with Existing Methods

EVM is based on extreme value theory, assuming that samples near class boundaries follow Weibull distributions. However, this method underperforms in this study, even slightly worse than the simple Softmax baseline. Analysis suggests that EVM's core assumptions may deviate from the actual characteristics of gas sensor data. Responses of gas-sensitive sensors are influenced by multiple physicochemical processes, and their feature distributions may exhibit atypical tail behaviors. Furthermore, EVM requires estimating boundary sample distributions in high-dimensional feature spaces, and relatively limited sample sizes may lead to inaccurate distribution estimation.

CAC effectively enhances intra-class feature compactness through the introduction of class anchor concepts and anchor constraint loss. However, CAC has two inherent limitations. It employs Euclidean distance as the scoring mechanism, which cannot adapt to anisotropic distributions of gas features. Additionally, it lacks dedicated drift handling mechanisms, meaning that when signal intensity changes, feature vector magnitudes change accordingly, causing Euclidean distance to produce deviations. SNM-Module simultaneously addresses both issues through geometric refinement and Mahalanobis distance.

### 3.7 Threshold Stability and Practical Value

Although this study employs adaptive thresholds in the experimental protocol to ensure fair comparison, the distribution overlap phenomenon in Figure 8(b) reveals an important finding: SNM-Module actually possesses the potential to use a globally fixed threshold. The core challenge faced by traditional methods is that rejection score distributions differ significantly under different conditions, causing thresholds determined under one condition to fail under others. SNM-Module eliminates this condition dependency through geometric refinement, meaning a single global threshold can be used to adapt to all positions, significantly reducing system deployment complexity and maintenance costs.

The high TPR performance achieved by SNM-Module has direct practical value in multiple industrial scenarios. In industrial safety monitoring, high TPR means the system misses virtually no potential threats, reducing miss rates to below 0.43%. In food quality control, high TPR ensures that substandard products are reliably identified and eliminated. In environmental monitoring, high TPR guarantees that regulatory authorities can promptly detect and respond to novel pollution events. The fixed 5% false positive rate constraint ensures system practicality, and SNM-Module achieves a 99.57% detection rate under this constraint, achieving an optimal balance between precision and practicality.

### 3.8 Limitations and Future Work

The training and testing data in this study originate from the same sensor array device, and the model's transfer capability across different devices has not yet been validated. Due to manufacturing variations and different aging states among sensor batches, cross-device deployment may face additional domain shift challenges. Future work will explore the combination of domain adaptation techniques with SNM-Module to enhance cross-device generalization capability. Additionally, although theoretical analysis indicates that SNM-Module incurs low computational overhead, systematic latency and power consumption tests on actual embedded platforms have not yet been conducted.

## 4. Conclusion

This study addresses the core challenges of open-set gas recognition in E-nose systems through an innovative geometric perspective. SNM-Module achieves feature-intensity decoupling through spherical normalization and constructs adaptive ellipsoidal decision boundaries through Mahalanobis distance, demonstrating three key innovations. The first is feature space geometric reconstruction that eliminates interference from signal intensity fluctuations at the physical root. The second is a statistical scoring mechanism that captures anisotropic feature distributions through covariance modeling. The third is an architecture-agnostic design that seamlessly enhances various backbone networks.

Systematic experiments on the Vergara gas sensor dataset validate the framework's effectiveness. Ablation experiments clearly demonstrate the individual contributions of each component, while cross-architecture validation demonstrates universal applicability. Compared to state-of-the-art methods, SNM-Module achieves a 3.02 percentage point improvement in AUROC and a 91.0% reduction in standard deviation, maintaining exceptional robustness across all five sensor positions. SNM-Module's unknown detection rate reaches 99.57%, and this breakthrough performance enables practical industrial deployment. This work establishes a new performance benchmark for drift-robust open-set gas recognition and provides solid theoretical and technical support for reliable machine olfaction systems in real industrial environments.

## Author Contributions

S.C.: Conceptualization, methodology, software, investigation, formal analysis, data curation, visualization, writing—original draft preparation. Y.S.: Formal analysis, validation, writing—review and editing. C.W.: Conceptualization, investigation, data curation. Z.W.: Conceptualization, methodology, resources, writing—review and editing, supervision, project administration, funding acquisition.

## Notes

The authors declare no competing financial interest.

## Acknowledgments

This work was supported by the National Natural Science Foundation of China (Grant No. 52305606), Joint Funds of the National Natural Science Foundation of China (Grant No. U25A20273), Key R&D Program of Shandong Province Shandong (Grant No. 2025JMRH0304), Provincial Natural Science Foundation (Grant No. ZR2023QE187), Basic Research Program of Jiangsu (Grant No. BK20230254), Qilu Young Scholars Program of Shandong University.